\documentclass[letterpaper, 10 pt, conference]{ieeeconf}  

\IEEEoverridecommandlockouts                              

\usepackage{cite}
\usepackage{graphicx} 
\usepackage{epsfig} 
\usepackage{times} 
\usepackage{amsmath} 
\usepackage{amssymb}  
\DeclareMathOperator{\atantwo}{atan2}
\usepackage{algorithm}
\usepackage{algorithmicx}
\usepackage[noend]{algpseudocode}
\usepackage{harpoon}
\usepackage{graphicx}
\usepackage{booktabs}  
\usepackage{threeparttable}
\usepackage{multirow}
\newcommand{\tabincell}[2]{\begin{tabular}{@{}#1@{}}#2\end{tabular}}
\usepackage{makecell}
\usepackage{color}
\usepackage{bbding}
\usepackage{stackengine}
\usepackage[hidelinks]{hyperref}

\title{\LARGE \bf
Long-Term Dynamic Window Approach for Kinodynamic Local Planning in Static and Crowd Environments
}

\author{Zhiqiang Jian$^{1}$, Songyi Zhang$^{1}$, Lingfeng Sun$^{2}$, Wei Zhan$^{2}$, Nanning Zheng$^{1\dag}$, and Masayoshi Tomizuka$^{2}$ 
\thanks{$^{1}$Z. Jian, S. Zhang, and N. Zheng are with the Institute of Artificial Intelligence and Robotics, Xi'an Jiaotong University, Xi'an, Shaanxi 710049, P.R. China; {\tt\small flztiii, zhangsongyi@stu.xjtu.edu.cn; nnzheng@mail.xjtu.edu.cn}}%
\thanks{$^{2}$L. Sun, W. Zhan, and M. Tomizuka are with the Department of Mechanical Engineering, University of California, Berkeley, CA 94720, USA; {\tt\small lingfengsun, wzhan, tomizuka@berkeley.edu}}%
\thanks{$^{\dag}$N. Zheng is the corresponding author.}%
}

\begin{document}

\maketitle
\thispagestyle{empty}
\pagestyle{empty}

\begin{abstract}

Local planning for a differential wheeled robot is designed to generate kinodynamic feasible actions that guide the robot to a goal position along the navigation path while avoiding obstacles. Reactive, predictive, and learning-based methods are widely used in local planning. However, few of them can fit static and crowd environments while satisfying kinodynamic constraints simultaneously. To solve this problem, we propose a novel local planning method. The method applies a long-term dynamic window approach to generate an initial trajectory and then optimizes it with graph optimization. The method can plan actions under the robot’s kinodynamic constraints in real time while allowing the generated actions to be safer and more jitterless. Experimental results show that the proposed method adapts well to crowd and static environments and outperforms most state-of-the-art approaches.

\end{abstract}

\section{INTRODUCTION}

Differential wheeled robot planning can be achieved by global and local planners. The global planner generates a navigation path to a goal point. The local planner continuously generates actions that guide the robot to follow the navigation path until it reaches the goal point. During this process, the actions generated by the local planner must meet the kinodynamic constraints and keep the robot safe and jitterless.

Tab.~\ref{tab:short_comings} shows several key features that should be satisfied for applying local planning methods in differential wheeled robots. First, planning methods need to obey the differential constraints and the acceleration limitation. Then, these methods should fit the static environments, which means they need to be able to deal with irregular borders. Meanwhile, they should also fit the crowd environments, which means they should be able to interact with multiple moving agents. Moreover, planning methods should be long-sighted, which means they need to give a long horizon planning result. Finally, planning methods should be able to track the navigation path to ensure the robot converges to the destination. However, as shown in Tab.~\ref{tab:short_comings}, most current local planning methods can only meet some of the above-mentioned features.

Therefore, we propose a novel local planning method satisfying all the features in Tab.~\ref{tab:short_comings}. First, our method constructs time-varying distance fields \cite{chen2018mobile} from the agents and occupancy grid map. Then, a Long-Term Dynamic Window Approach (LT-DWA) is proposed to generate a long-time horizon state-cost tree. Finally, a path with the least cost in the tree is selected and optimized using the Elastic-Band Model Predictive Control (EB-MPC) method to obtain the planned trajectory.

\begin{table}[!t]
    \centering
    \renewcommand{\arraystretch}{1.0}
    \fontsize{8}{8}\selectfont
    \begin{threeparttable}
        \caption{Features for local planning methods applied in the differential mobile robots.} 
        \label{tab:short_comings}
        \setlength{\tabcolsep}{3pt}
        \begin{tabular}{ccccccc}
        \toprule
        Method & \tabincell{c}{Differential\\Constraints} & \tabincell{c}{Acc.\\Limit} & \tabincell{c}{Static\\Env.} & \tabincell{c}{Crowd\\Env.} & \tabincell{c}{Long\\Sighted} & \tabincell{c}{Track\\Nav.} \cr
        \midrule
        \midrule
        DWA \cite{fox1997dynamic} & \Checkmark & \Checkmark & \Checkmark & \XSolid & \XSolid & \XSolid \cr
        PCL-LSTM \cite{leiva2020robust} & \Checkmark & \Checkmark & \Checkmark & \XSolid & \Checkmark & \XSolid \cr
        ESA \cite{shi2022enhanced} & \Checkmark & \XSolid & \XSolid & \Checkmark & \Checkmark & \XSolid \cr
        SOADRL \cite{liu2020robot} & \XSolid & \XSolid & \Checkmark & \Checkmark & \Checkmark & \XSolid \cr
        TEB \cite{rosmann2017kinodynamic} & \Checkmark & \Checkmark & \Checkmark & \XSolid & \Checkmark & \Checkmark \cr
        KCP \cite{shin2018kinodynamic} & \Checkmark & \Checkmark & \Checkmark & \XSolid & \Checkmark & \XSolid \cr
        DC \cite{cao2019dynamic} & \Checkmark & \XSolid & \XSolid & \Checkmark & \Checkmark & \XSolid \cr
        Timed-ESDF \cite{zhu2022online} & \XSolid & \XSolid & \Checkmark & \Checkmark & \Checkmark & \XSolid \cr
        Our method & \Checkmark & \Checkmark & \Checkmark & \Checkmark & \Checkmark & \Checkmark \cr
        \bottomrule
        \end{tabular}
    \end{threeparttable}
\end{table}

In conclusion, this paper has the following contributions:
\begin{itemize}
    \item The LT-DWA is proposed to generate the initial state sequence.
    \item Time-varying distance fields are combined with the MPC \cite{brito2019model} to formulate the planning problem and the EB method \cite{quinlan1993elastic} is applied to solve it.
    \item The proposed local planner is open-sourced\footnote{\url{https://github.com/flztiii/LT_DWA}}. It can be applied to static and crowd environments and outperforms current planning methods.
\end{itemize}

\section{RELATED WORK}

The local planner can be achieved by reactive, predictive, and learning-based methods. Reactive methods directly build the mapping from the robot's current state to action, including Dynamic Window Approach (DWA) \cite{fox1997dynamic}, Reciprocal Velocity Obstacles (RVO) \cite{van2008reciprocal}, and RouteGAN\cite{sun2021diverse}. Predictive methods generate a continuous sequence of actions based on the robot's current state and predicted future conditions. For example, the Timed Elastic Band (TEB) method proposed by Rösmann \emph{et al.} \cite{rosmann2017kinodynamic}, the Dynamic Channel (DC) method proposed by Cao \emph{et al.} \cite{cao2019dynamic}, and the Timed-ESDF method proposed by Zhu \emph{et al.} \cite{zhu2022online} are all predictive methods. Learning-based methods use large amounts of data to map from the robot's current state to its action through imitation learning or reinforcement learning. Learning-based methods such as SARL \cite{chen2019crowd}, RGL \cite{chen2020relational}, DSRNN \cite{liu2021decentralized}, and ESA \cite{shi2022enhanced} show state-of-the-art performance in crowd environments by considering the crowd's interaction \cite{alahi2016social, sun2023distributed}.

\textbf{MPC-formed planning:} Modeling the local planning problem in MPC form and then solving the problem through optimization is an effective approach \cite{lam2010model, brito2019model, pauls2022real}. The mathematical models established by these methods are often non-convex, so the initial guess significantly influences the planning result. However, obtaining a feasible initial guess in a crowd environment is challenging. For example, Brito \emph{et al.}'s method \cite{brito2019model} uses the expansion of the previous planning result as the initial guess for the next planning episode, which is difficult to guarantee the feasibility of the initial guess in a crowd environment. Therefore, in our method, we introduce a robust method for generating a feasible initial guess and combine it with the MPC-formed planning method.

\textbf{DWA methods:} The DWA has many applications as a planning method that considers kinodynamic and environmental constraints simultaneously. Its fundamental idea is to sample in the feasible control space and then evaluate in the state space. Brock \emph{et al.} \cite{brock1999high} improve the DWA's evaluation function. Ogren \emph{et al.} \cite{ogren2002tractable} combine the DWA with global planning and prove the global convergence of their method. However, none of their methods can address the short-sight of the DWA in a single planning episode. A solution is to use multi-step DWA, but another problem will arise: the exponential expansion of the state space. If the exponential expansion of the state space of the multi-step DWA can be solved, it can be applied to generating the feasible initial guess, which our method does.

\textbf{Distance field:} The distance field adequately represents the environment and is widely used by various planners \cite{oleynikova2016signed, chen2018mobile, ngo2020develop}. Oleynikova \emph{et al.}'s method \cite{oleynikova2016signed} introduces the distance field in the static environment. Chen \emph{et al.}'s and Ngo \emph{et al.}'s methods \cite{chen2018mobile, ngo2020develop} are proposed to construct the distance field in the crowd environment and prove its effectiveness. Therefore, we introduce the distance field into the MPC-formed planning method to improve the effectiveness of the planner.

\section{PROBLEM FORMULATION}

The state and system dynamics of the robot are defined as follows.
\begin{equation}
    \label{eq:robot_define}
    \begin{aligned}
        &\mathbf{s}=(x, y, \theta, v, \omega)^T, \quad \mathbf{u}=(a_v, a_\omega)^T, \\
        &\dot{\mathbf{s}} = f(t, \mathbf{s}, \mathbf{u}) = (v\cos{\theta}, v\sin{\theta}, \omega, a_v, a_\omega)^T, \\
    \end{aligned}
\end{equation}
where, $x$ and $y$ indicate the position in 2D space, $\theta$ is orientation, $v$ and $\omega$ are linear and angular velocities, $a_v$ and $a_\omega$ are linear and angular accelerations.

The robot's radius is defined as $R$. The robot's state at time $t_0$ is defined as $\mathbf{s}_\mathrm{init}$. Since all the inputs can be transformed into the coordinate system with the robot state as the origin, without losing generality, it can be considered that $\mathbf{s}_\mathrm{init}=\mathbf{0}^T \times v_\mathrm{init} \times w_\mathrm{init}$. The navigation path connecting the robot's current and goal points is defined as $\mathcal{P}=\{\mathbf{p}_p\}_{p=0}^P$, where $\mathbf{p}_p=(x_{\mathbf{p}_p}, y_{\mathbf{p}_p}, \theta_{\mathbf{p}_p})^T$ indicates the $p_\mathrm{th}$ point's position and orientation on the navigation path. The agents are defined as $\mathcal{O}=\{\mathbf{o}_o\}_{o=0}^O$, where $\mathbf{o}_o=(x_{\mathbf{o}_o}, y_{\mathbf{o}_o}, vx_{\mathbf{o}_o}, vy_{\mathbf{o}_o}, r_{\mathbf{o}_o})^T$ indicates the $o_\mathrm{th}$ agent's center position, velocity, and radius. The occupancy grid map is defined as $\mathcal{B}=\{\mathbf{b}_b\}_{b=0}^B$, where $\mathbf{b}_b=(x_{\mathbf{b}_b}, y_{\mathbf{b}_b})^T$ indicates the $b_\mathrm{th}$ occupied grid's position.

In this way, the local planning problem can be described as follows. The local planning is to calculate the state sequence $\mathbf{s}(t), t \in [t_0, t_0+T]$ within a fixed time $T$, so that $\mathbf{s}(t)$ minimizes the cost function $c(\mathbf{s}(t), \mathcal{P}, \mathcal{O}, \mathcal{B}, T)$ while satisfying Eq.~\ref{eq:robot_define}. For simplification, the problem is discretized in the time domain, where $T$ is divided into $N$ frames time segments $\Delta T$. Then, $\mathbf{s}(t)$ can be defined as $\mathcal{S}=\{\mathbf{s}_i\}_{i=0}^N$ ($\mathbf{s}_i=\mathbf{s}(t_0 + i \Delta T)$), and the local planning problem can be defined as follows.
\begin{equation}
    \label{eq:local_planning_problem}
    \begin{aligned}
        &\min \limits_{\mathcal{S}} \quad c(\mathcal{S}, \mathcal{P}, \mathcal{O}, \mathcal{B}) \\
        &\mathrm{s.t.} \quad \left\{
        \begin{array}{l}
            \mathbf{s}_0 = \mathbf{0}^T \times v_\mathrm{init} \times w_\mathrm{init}, \\
            \mathbf{s}_{i+1} = \mathbf{s}_i + \Delta T f(t_0 + i \Delta T, \mathbf{s}, \mathbf{u}), \\
            ~~~~~~~~~~\forall i \in [0, N-1], \\
            \mathbf{s}_i \in \mathrm{SE}(2) \times [v_\mathrm{min}, v_\mathrm{max}] \times [\omega_\mathrm{min}, \omega_\mathrm{max}], \\
            \mathbf{u}_i \in [a_v^\mathrm{min}, a_v^\mathrm{max}] \times [a_\omega^\mathrm{min}, a_\omega^\mathrm{max}],
        \end{array}
        \right.
    \end{aligned}
\end{equation}
where, function $f(\cdot)$ is the same as Eq.~\ref{eq:robot_define}, $v_\mathrm{min}$ and $v_\mathrm{max}$ are the robot's minimum and maximum linear velocities, $\omega_\mathrm{min}$ and $\omega_\mathrm{max}$ are the robot's minimum and maximum angular velocities, $a_v^\mathrm{min}$ and $a_v^\mathrm{max}$ are the robot's minimum and maximum linear accelerations, $a_\omega^\mathrm{min}$ and $a_\omega^\mathrm{max}$ are the robot's minimum and maximum angular accelerations. In our method, the cost function $c(\cdot)$ is the weighted summation of the collision risk cost $c_\mathrm{c}(\cdot)$, navigation following cost $c_\mathrm{n}(\cdot)$, and jitter cost $c_\mathrm{j}(\cdot)$, which are defined in section IV. The collision risk cost aims to improve the robot's safety, which means a lower collision rate and a longer distance to the borders. The navigation following cost punishes the robot from deviating from the navigation path. The jitter cost aims to reduce the robot's jitter, which means a smaller change in the orientation, linear velocity, and angular velocity.

\section{METHOD}

\subsection{Framework}

\begin{figure*}[!t]
    \centering
    \includegraphics[width=0.96\linewidth]{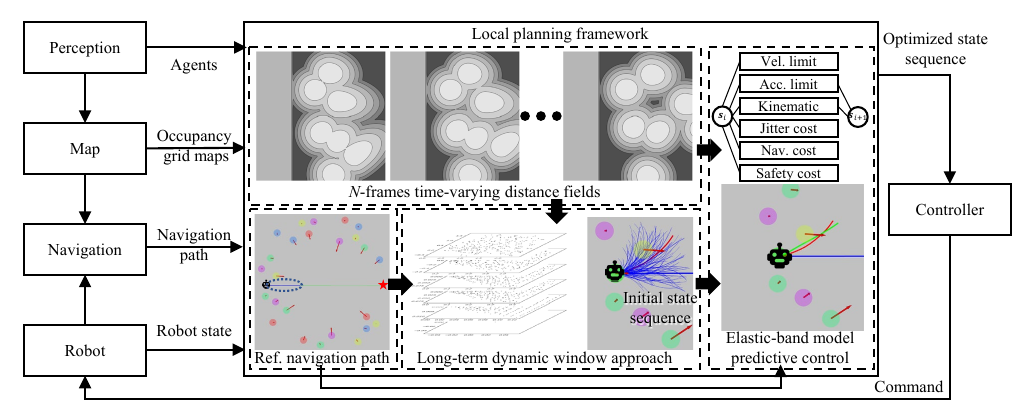}
    \caption{This figure shows the local planner's framework. The example of the time-varying distance fields is shown in the upper left. The example of the reference navigation path is shown in the bottom left, where the green line indicates the navigation path, the blue line indicates the reference navigation path, the colorful circles indicate agents, and the red arrows indicate the agents' velocities. The example of the LT-DWA is shown at the bottom, where the left side indicates the expanded states in the different frames, the blue curves on the right side indicate the projection of the state-cost tree on the Cartesian plane, and the red curve indicates the selected state sequence. The example of EB-MPC is shown on the right, where the upper part indicates the connection relationship of a node in the graph optimization, and the green curve in the lower part indicates the optimized state sequence.}
    \label{fig:framework}
\end{figure*}

The framework shown in Fig.~\ref{fig:framework} is proposed to solve the local planning problem. In step one (Section B), the local planner generates the $N$-frames time-varying distance fields $\{d_i(\cdot)\}_{i=0}^N$. ($d_i(\cdot): \mathbb{R}^2 \! \rightarrow \! \mathbb{R}$ is a mapping from the Cartesian plane to the real number set.)  Then, the planner calculates the reference navigation path $\{\mathbf{p}_i\}_{i=0}^N$. ($\mathbf{p}_i=(x_{\mathbf{p}_i}, y_{\mathbf{p}_i}, \theta_{\mathbf{p}_i})^T \in \mathrm{SE}(2)$.) $\{\mathbf{p}_i\}_{i=0}^N$ and $\{d_i(\cdot)\}_{i=0}^N$ are required in the following state cost and optimization objective function calculations. In step two (Section C), the planner applies LT-DWA to generate an initial state sequence. In step three (Section D), the planner uses the EB-MPC method to optimize the initial state sequence. According to the optimized state sequence, the controller generates control commands. The local planner updates at a fixed frequency until the robot reaches the goal point.

\subsection{Reference Navigation Path and Time-Varying Distance Fields}

\begin{figure}[!t]
    \centering
    \includegraphics[width=0.95\linewidth]{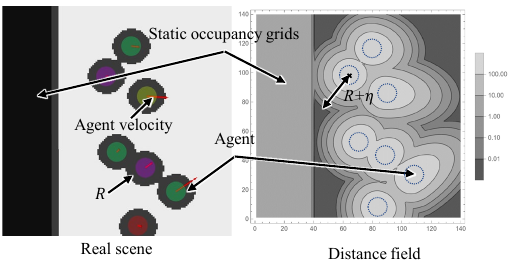}
    \caption{In this figure, the left part shows the scenario and the right part shows the first frame of its corresponding time-varying distance fields. The colorful circles indicate agents, the red arrows indicate agents' velocities, and the black region indicates the occupancy grid map.}
    \label{fig:distance_field}
\end{figure}

The reference navigation path $\{\mathbf{p}_i\}_{i=0}^N$ is obtained according to the navigation path $\mathcal{P}$. The point on the navigation path, which has the minimum euclidean distance to the robot's current state $\mathbf{0}^T$ on the Cartesian plane, is selected as the reference navigation path's beginning point $\mathbf{p}_0$. Then, the $i_\mathrm{th}$ point $\mathbf{p}_i$ on the reference navigation path is the point on the navigation path, whose arc length to $\mathbf{p}_0$ along the navigation path is $i \Delta T v_\mathrm{max} \max(\cos \theta_{\mathbf{p}_0}, 0)$. $\theta_{\mathbf{p}_0}$ is the orientation gap between $\mathbf{p}_0$ and $\mathbf{0}^T$. This expression expects that when the difference between the orientation of the robot and the navigation path is significant, the robot should adjust its orientation first. Otherwise, it should follow the navigation path at the maximum speed.

The time-varying distance fields $\{d_i(\cdot)\}_{i=0}^N$ are calculated according to the agents $\mathcal{O}$ and the occupancy grid map $\mathcal{B}$. The $i_\mathrm{th}$ frame distance field $d_i(\cdot)$ is determined by the agent distance field $d_i^\mathcal{O}(\cdot)$ and the occupancy grid map distance field $d_i^\mathcal{B}(\cdot)$, whose calculation is as follows.
\begin{equation}
    \label{eq:distance_field}
    d_i(\cdot) = \max(w_\mathrm{do} d_i^\mathcal{O}(\cdot), w_\mathrm{db} d_i^\mathcal{B}(\cdot)),
    \notag
\end{equation}
where, $w_\mathrm{do}$ and $w_\mathrm{db}$ are preset weights.

$d_i^\mathcal{O}(\cdot)$ is calculated as follows, referring to Chen \emph{et al.}'s method \cite{chen2018mobile}.
\begin{equation}
    \label{eq:distance_field_o}
    d_i^\mathcal{O}(x, y) \!=\! \max_{\forall o \in [0, O]} \exp^{-\big{(}\frac{l_x^o(x, y, i)^2}{2\sigma_x^2} + \frac{l_y^o(x, y, i)^2}{2\sigma_y^2}\big{)}},
    \notag
\end{equation}
where,
\begin{equation}
    \begin{aligned}
         &l^o(x, y, i) = \Big \| \begin{pmatrix} x \\ y \end{pmatrix} - \begin{pmatrix} x_{\mathbf{o}_o} + i \Delta T vx_{\mathbf{o}_o} \\ y_{\mathbf{o}_o} + i \Delta T vy_{\mathbf{o}_o} \end{pmatrix} \Big \|, \\
         &\alpha = \atantwo \Big (\frac{y \!-\! y_{\mathbf{o}_o} \!-\! i \Delta T vy_{\mathbf{o}_o}}{x \!-\! x_{\mathbf{o}_o} \!-\! i \Delta T vx_{\mathbf{o}_o}} \Big ) - \atantwo \Big (\frac{vy_{\mathbf{o}_o}}{vx_{\mathbf{o}_o}} \Big ), \\
         &l_x^o(x, y, i) \!=\! l^o(x, y, i) \!*\! \cos \alpha, l_y^o(x, y, i) \!=\! l^o(x, y, i) \!*\! \sin \alpha, \\
         &\sigma_x = \left\{
            \begin{array}{l}
            \!\frac{1}{3}\! \big (r_{\mathbf{o}_o} \!+\! R \!+\! \eta \!+\! \beta \big \|\begin{pmatrix} vx_{\mathbf{o}_o} \\ vy_{\mathbf{o}_o} \end{pmatrix} \big \| \big ), \quad -\frac{\pi}{2} \!<\! \alpha \!<\! \frac{\pi}{2}, \\
            \!\frac{1}{3}\! (r_{\mathbf{o}_o} \!+\! R \!+\! \eta), \quad \mathrm{else}, \\
            \end{array}
         \right. \\
         &\sigma_y = \frac{1}{3} (r_{\mathbf{o}_o} + R + \eta),
    \end{aligned}
    \notag
\end{equation}
where, $\eta$ and $\beta$ are preset parameters. As shown in Fig.~\ref{fig:distance_field}, this distance field comprises multiple two-dimensional normal distributions with offset. The normal distributions' centers are $i_\mathrm{th}$ frame agents' centers, whose ranges are determined by $\eta$ and offsets are determined by the agents' velocities and $\beta$.

$d_i^\mathcal{B}(\cdot)$ is calculated as follows.
\begin{equation}
    \begin{aligned}
        &d_i^\mathcal{B}(x, y) = (\mathrm{Relu}(\eta - \delta ))^2, \\
        & \delta = \min_b \mathrm{Relu} \Big ( \big \| \begin{pmatrix} x - x_{\mathbf{b}_b} \\ y - y_{\mathbf{b}_b} \end{pmatrix} \big \| - R \Big ), \quad \forall b \in [0, B],
    \end{aligned}
    \notag
\end{equation}
where, $\eta$ is the same preset parameter. As shown in Fig~\ref{fig:distance_field}, this distance field is a quadratically decreasing function with the distance to the boundary of the occupancy grid map. The value of this distance field is zero, if the distance to the boundary of the occupancy grid map is larger than $\eta+R$.

\subsection{Long-Term Dynamic Window Approach}

After obtaining the reference navigation path $\{\mathbf{p}\}_{i=0}^N$ and the time-varying distance fields $\{d_i(\cdot)\}_{i=0}^N$, the next step is to build a state-cost tree $\mathcal{T}$ to obtain the initial state sequence $\mathcal{S}_\mathrm{init}$ for the following optimization. The tree is a set of $N$ layers, and the tree's $i_\mathrm{th}$ layer $\mathcal{T}_i$ is the set of $i_\mathrm{th}$ frame nodes. Each node $n$ has three attributes: state (State$[n]$), cost (Cost$[n]$), and parent node (Parent$[n]$). The construction of the state-cost tree is shown in Alg.~\ref{alg:lt_dwa}.

\begin{algorithm}[!t]  
	\caption{Long Term Dynamic Window Approach}  
	\label{alg:lt_dwa}  
	\begin{algorithmic}[1]
		\Require Reference Navigation Path $\{\mathbf{p}\}_{i=0}^N$, Time-Varying Distance Fields $\{d_i(\cdot)\}_{i=0}^N$.
		\Ensure State-Cost Tree $\mathcal{T}$.
        \State  Add Layer $\mathcal{T}_0=\{$Node($\mathbf{0}^T$, 0, null)$\}$ into empty Tree $\mathcal{T}$.
		\For{$i=1,2,\cdots,N$}
            \State $\mathcal{T}_i$ $\leftarrow$ $\varnothing$.
            \For{Node $n$ $\in$ Layer $\mathcal{T}_{i-1}$}
                \State $\mathcal{S}_\mathrm{c}$ $\leftarrow$ expandStates($n$).
                \For{State $\mathbf{s}$ $\in$ $\mathcal{S}_\mathrm{c}$}
                    \State Push Node($\mathbf{s}$, 0, $n$) to Layer $\mathcal{T}_i$.
                \EndFor
            \EndFor
            \If{len($\mathcal{T}_{i}$) $>$ $K^{'}$}
                \State $\mathcal{T}_{i}$ $\leftarrow$ voxelSampling($\mathcal{T}_{i}$).
            \EndIf
            \For{Node $n$ $\in$ Layer $\mathcal{T}_{i}$}
                \State Cost$[n]$ $=$ Cost$[$Parent$[n]$$] +$ calcCost(State$[n]$, $\mathbf{p}_i$, $d_i(\cdot)$, $i$).
            \EndFor
            \If{$\mathcal{T}_i$ is not $\varnothing$}
                \State Add Layer $\mathcal{T}_i$ into Tree $\mathcal{T}$.
            \Else
                \State \textbf{break}
            \EndIf
        \EndFor
	\end{algorithmic}  
\end{algorithm}

\begin{figure}[!t]
    \centering
    \includegraphics[width=\linewidth]{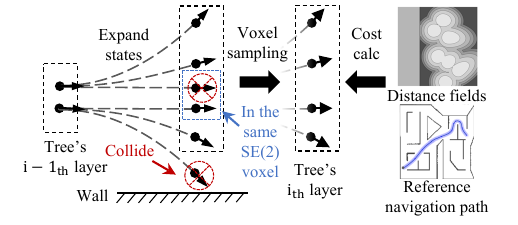}
    \caption{This figure shows an exemplary diagram of how to construct the layer of the tree. The diagram includes state expansion, voxel sampling, and cost calculation.}
    \label{fig:tree_build}
\end{figure}

The root layer $\mathcal{T}_0$ can be obtained according to the robot's current state. Then, $\mathcal{T}_i$ is obtained as the exemplary diagram in Fig.~\ref{fig:tree_build}. All the nodes in the previous layer $\mathcal{T}_{i-1}$ are traversed. For each node $n$, state expansion is performed according to $n$'s state State$[n]$. In this way, an expanded state set $\mathcal{S}_\mathrm{c}$ can be generated. The DWA achieves state expansion. In detail, given a state $\mathbf{s}$, the velocity space boundary of the states in the next frame can be determined according to the robot's velocity and acceleration limitations. The limited velocity space is uniformly sampled, and $V \times V$ samples can be obtained. For each sample, an expanded state can be calculated according to Eq.~\ref{eq:robot_define}. Collision-free states are selected from these expanded states, which make up $\mathcal{S} _\mathrm{c}$. A new node without cost is added to $\mathcal{T}_i$ for each state in $\mathcal{S}_\mathrm{c}$.

When the states of all nodes in the previous frame are expanded, the number of nodes in $\mathcal{T}_{i}$ is significant. Assuming that the DWA can expand $K$ states from one state each time, the time complexity of calculating the tree with $N$ frames is $O (K^N)$. When $N$ is large, this time complexity is undoubtedly unacceptable. In order to solve this problem, after using the DWA to expand all the states in the current frame, we perform voxel sampling to ensure that the total number of states in the next frame is always around $K^{'}$. In this way, the time complexity is $O(K^{'}N)$. The voxel sampling process is as follows. The $\mathrm{SE}(2)$ space boundary of all nodes in $\mathcal{T}_{i}$ is calculated, and subsequently, the $\mathrm{SE}(2)$ space within the boundary is voxelized to $W \times W \times W$ voxels. Each node is located in one voxel. A node is randomly sampled in each voxel, and the sampling result can be obtained. 

\begin{figure}[!t]
    \centering
    \includegraphics{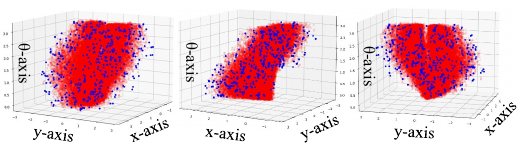}
    \caption{This figure shows an example of the robot's state distribution and the sampled state distribution in the SE(2) space at the end of the long horizon. The red points represent the robot's state distribution, and the blue points represent the sampled state distribution using voxel sampling.}
    \label{fig:distribution}
    \vspace{-0.2in}
\end{figure}

There are two reasons why voxelization is performed in the $\mathrm{SE}(2)$ space instead of the state space. The first is to reduce the space dimension and achieve lower computational complexity. The second is that the attributes of the state in the $\mathrm{SE}(2)$ space are more important than those in the velocity space. The former has a direct relationship to the robot's safety, while the latter only has an indirect relationship. Fig.~\ref{fig:distribution} shows an example of the robot's state distribution and the sampled state distribution in the $\mathrm{SE}(2)$ space at the end of the long horizon. It can be seen from the figure that the blue points distribution is consistent with the red points distribution. Therefore, the representation of the robot's $\mathrm{SE}(2)$ properties at the horizon's end can be achieved using voxel sampling.

After voxel sampling, the number of nodes in $\mathcal{T}_{i}$ will be acceptable for real-time performance. At this time, the cost of each node $n$ in $\mathcal{T}_{i}$ can be calculated, which includes two parts. One is the cost of its parent node, and the other is its state $\mathbf{s}$'s cost. When calculating the cost of $\mathbf{s}$, only the collision risk cost $c_\mathrm{c}(\cdot)$ and the navigation following cost $c_\mathrm{n}(\cdot)$ are considered. The reason is that $c_\mathrm{n}(\cdot)$ and $c_\mathrm{c}(\cdot)$ are related to $\mathrm{s}$'s $\mathrm{SE}(2)$ space attributes, while the jitter cost $c_\mathrm{j}(\cdot)$ is only related to the velocity space attributes. In the previous step, the sampling is performed in the $\mathrm{SE}(2)$ space, so the calculation of $c_\mathrm{j}(\cdot)$ does not make much sense here. In conclusion, given the $i_\mathrm{th}$ frame point on the reference navigation path $\mathbf{p}_i=(x_{\mathbf{p}_i}, y_{\mathbf{p}_i}, \theta_{\mathbf{p}_i})^T$ and the $i_\mathrm{th}$ frame time-varying distance field $d_i(\cdot)$, the cost of the $i_\mathrm{th}$ frame state $\mathbf{s}=(x_\mathbf{s}, y_\mathbf{s}, \theta_\mathbf{s}, v_\mathbf{s}, \omega_\mathbf{s})^T$ can be calculated as follows.
\begin{equation}
    \label{eq:state_cost}
    \begin{aligned}
        &\mathrm{calcCost}(\mathbf{s}, \mathbf{p}_i, d_i(\cdot), i)\!=\! \gamma^i (c_\mathrm{c}(\mathbf{s}, d_i(\cdot)) + c_\mathrm{n}(\mathbf{s}, \mathbf{p}_i)), \\
        &c_\mathrm{c}(\mathbf{s}, d_i(\cdot))) = w_\mathrm{c} d_i(x_\mathbf{s}, y_\mathbf{s}), \\
        &c_\mathrm{n}(\mathbf{s},\! \mathbf{p}_i) \!=\! w_\mathrm{no} c_\mathrm{no}(\mathbf{s},\! \mathbf{p}_i) \!\!+\!\! w_\mathrm{na} c_\mathrm{na}(\mathbf{s},\! \mathbf{p}_i) \!\!+\!\! w_\mathrm{nt} c_\mathrm{nt}(\mathbf{s},\! \mathbf{p}_i),
    \end{aligned}
\end{equation}
where,
\begin{equation}
    \label{eq:navigation_cost}
    \begin{aligned}
        &c_\mathrm{no}(\mathbf{s}, \mathbf{p}_i) \!=\! ((x_\mathbf{s} \!-\! x_{\mathbf{p}_i})\cos{\theta_{\mathbf{p}_i}} \!+\! (y_\mathbf{s} \!-\! y_{\mathbf{p}_i})\sin{\theta_{\mathbf{p}_i}})^2, \\
        &c_\mathrm{na}(\mathbf{s}, \mathbf{p}_i) \!=\! (\!-(x_\mathbf{s} \!-\! x_{\mathbf{p}_i})\sin{\theta_{\mathbf{p}_i}} \!+\! (y_\mathbf{s} \!-\! y_{\mathbf{p}_i})\cos{\theta_{\mathbf{p}_i}})^2, \\
        &c_\mathrm{nt}(\mathbf{s}, \mathbf{p}_i) = (1 - \cos(\theta_\mathbf{s} - \theta_{\mathbf{p}_i}))^2, \\
    \end{aligned}
\end{equation}
$\gamma$ is decline rate, $w_\mathrm{c}$, $w_\mathrm{no}$, $w_\mathrm{na}$, and $w_\mathrm{nt}$ are preset weights. The reason for setting the decline rate is uncertainty in the movement of agents in the crowd environment, and the cost in the long-term future has low reliability. The cost $c_\mathrm{no}(\cdot)$ is to penalize the longitudinal distance between $\mathbf{s}$ and $\mathbf{p}_i$ and the cost $c_\mathrm{na}(\cdot)$ is to penalize the lateral distance between $\mathbf{s}$ and $\mathbf{p}_i$. The combination of the two costs can be used to evaluate the distance between $\mathbf{s}$ and the reference navigation path in the Cartesian plane, which is more flexible than directly calculating the distance between $\mathbf{s}$ and the reference navigation path \cite{abbas2017obstacle}. The cost $c_\mathrm{nt}(\cdot)$ is to penalize the orientation gap between $\mathbf{s}$ and $\mathbf{p}_i$.

After $\mathcal{T}$ is built, the nodes in the $N_\mathrm{th}$ layer $\mathcal{T}_N$ of the tree are traversed, and the node with the minimum cost is selected. We iteratively backtrack the node's parent until the tree's root node is reached. Finally, the initial state sequence $\mathcal{S}_\mathrm{init}$ can be obtained. In the complex environment, when the $N^{'}_\mathrm{th}$ layer of the tree is built, it may be empty sometimes. In this case, building a complete $N$ layer tree is given up and a $N^{'}\!-\!1$ layer tree is obtained. Then, $\mathcal{S}_\mathrm{init}$ also degenerates from $N$ frames to $N^{'}\!-\!1$ frames. In the worst case, this method will degenerate into the DWA.

\subsection{Elastic-Band Model Predictive Control}

After obtaining the initial state sequence $\mathcal{S}_\mathrm{init}$, the next step is to optimize it using the EB-MPC method. We define the state sequence optimization problem in the MPC form \cite{brito2019model} and solve the problem using the EB method \cite{quinlan1993elastic}.

The optimization model used has been given in Eq.~\ref{eq:local_planning_problem}, which is an MPC-formed model. The expression of the objective function $c(\mathcal{S}, \mathcal{P}, \mathcal{O}, \mathcal{B})$ is as follows.
\begin{equation}
    \label{eq:objective_function}
    \begin{aligned}
        c(\mathcal{S},\! \mathcal{P},\! \mathcal{O},\! \mathcal{B}) \!\!=\!\! \sum_{i=1}^N \gamma^i (&c_\mathrm{c}(\mathbf{s}_i, d_i(\cdot)) \!\!+\!\! c_\mathrm{n}(\mathbf{s}_i,\! \mathbf{p}_i) \!\!+\!\! c_\mathrm{j}(\mathbf{s}_i,\! \mathbf{s}_{i-1})),
    \end{aligned}
    \notag
\end{equation}
where, $\gamma$ and $c_\mathrm{c}(\cdot)$ is the same as Eq.~\ref{eq:state_cost}. $c_\mathrm{n}(\cdot)$ adds an additional cost $c_\mathrm{nv}(\cdot)$ on the basis of Eq.~\ref{eq:state_cost}, whose expression is as follows.
\begin{equation}
    c_\mathrm{nv}(\mathbf{s}, \mathbf{p}_i) = w_\mathrm{nv} (v_\mathbf{s} - \min(v_\mathrm{max}, \sqrt{2 a_\mathrm{max} \epsilon_{\mathbf{p}_i}}))^2,
    \notag
\end{equation}
where, $w_\mathrm{nv}$ is preset weight, $v_\mathbf{s}$ is $\mathbf{s}$'s linear velocity, $\epsilon_{\mathbf{p}_i}$ is arc length between $\mathbf{p}_i$ and the navigation path's endpoint. This expression aims to make the robot's speed tend to the maximum value when it is far away from the goal point. Otherwise, the speed tends to zero. $c_\mathrm{j}(\cdot)$ is the jitter cost, whose expression is as follows.
\begin{equation}
    \label{eq:jitter_cost}
    \begin{aligned}
    c_\mathrm{j}(\mathbf{s}_i,\! \mathbf{s}_{i-1}) \!=\! &w_\omega \omega_i^2 \!\!+\!\! w_{a_v} \Big{(}\frac{v_i \!\!-\!\! v_{i-1}}{\Delta T}\Big{)}^2 \!\!+\!\! w_{a_\omega} \Big{(}\frac{w_i \!\!-\!\! w_{i-1}}{\Delta T}\Big{)}^2,
    \end{aligned}
    \notag
\end{equation}
where, $w_\omega$, $w_{a_v}$, and $w_{a_\omega}$ are preset weights, $\mathbf{s}_i=(x_i, y_i, \theta_i, v_i, \omega_i)^T$. There are three items in the jitter cost. The first item is to penalize high angular velocities for reducing the shaking of the robot's orientation. The second item is to penalize high linear accelerations for reducing the jitter of the robot's speed. The third item is to penalize high angular accelerations for reducing the vibration of the robot's angular velocity. The above three items work together to reduce the jitter of the robot.

After obtaining the complete definition of the optimization model, it can be seen that the optimization model Eq.~\ref{eq:local_planning_problem} is sparse for the optimization variable $\mathcal{S}$, so the EB method can be used to solve it \cite{rosmann2013efficient}, whose process is as follows. Each state $\mathbf{s}_i$ in the optimization variable $\mathcal{S}$ can be regarded as a node, and the objective function and constraints in the optimization model Eq.~\ref{eq:local_planning_problem} can be regarded as edges. In this way, a graph can be constructed, as shown on the right side of Fig.~\ref{fig:framework}. According to Eq.~\ref{eq:local_planning_problem}, there are only unary edges and binary edges in the constructed graph. Subsequently, $\mathcal{S}_\mathrm{init}$ is applied to initialize the graph, and the g2o framework \cite{grisetti2011g2o} is used to perform graph optimization. The algorithm used for graph optimization is the Levenberg-Marquardt. At last, the optimized state sequence $\mathcal{S}_\mathrm{opt}$ can be obtained. Compared with $\mathcal{S}_\mathrm{init}$, $\mathcal{S}_\mathrm{opt}$ can make the robot have less jitter.

\begin{figure*}[htbp]
    \centering
    \includegraphics[width=\linewidth]{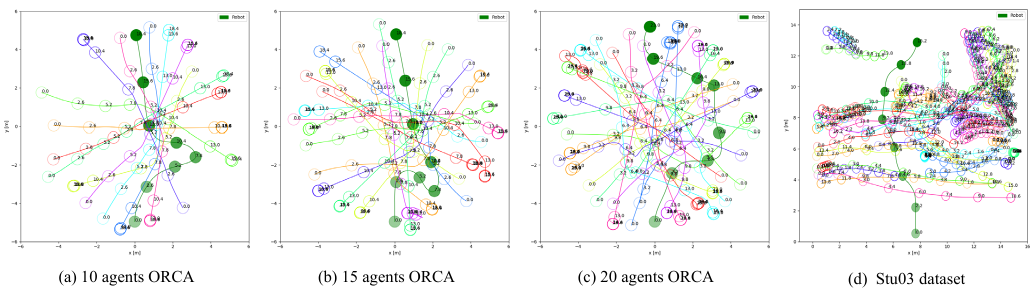}
    \vspace{-0.2in}
    \caption{This figure shows four examples of the robot navigating in different crowds using the proposed method. The dark green circles indicate the robot, the colorful hollow circles indicate other agents, the curves of different colors indicate the moving trajectories of the corresponding agents, and the values near the circles indicate the corresponding time. The more transparent the circle, the earlier the time.}
    \vspace{-0.2in}
    \label{fig:crowd_test}
\end{figure*}

\section{EXPERIMENTAL RESULTS}

The experiments are conducted in crowd, static, and hybrid environments to verify our method in different scenarios. In addition, we design an ablation study to verify the effectiveness of submodules. The testing robot is designed as a differential wheel robot. The robot's shape is set as a 0.3 m radius circle, and its sensing range is limited to 3.5 m. The robot's linear velocity is set from 0 to 1 m/s, its angular velocity is set from -1 rad/s to 1 rad/s, its linear acceleration is set from -1 m/$\mathrm{s}^2$ to 1 m/$\mathrm{s}^2$, and its angular acceleration is set from -1 rad/$\mathrm{s}^2$ to 1 rad/ $\mathrm{s}^2$.

\subsection{Crowd Environment Tests}

In this experiment, the ORCA simulated scenarios as the ESA \cite{shi2022enhanced} and the pedestrian trajectory dataset \cite{lerner2007crowds} are both used for testing. The testing scenarios are updated at a frequency of 5 Hz.

For the ORCA simulated scenarios, 10, 15, and 20 agents are set in the environment, respectively. Each agent is a circle with a radius of 0.3 m, its moving policy is the ORCA, and its maximum speed is 1 m/s. The robot is invisible to all the agents \cite{chen2019crowd, chen2020relational, shi2022enhanced}. In each test, the agents are on a circle with a radius of 5 m and moves to the targets, which are their opposite position of the circle with disturbance. Meanwhile, the robot is also on the circle and regards the opposite position as the goal point, as shown in Fig.~\ref{fig:crowd_test}.

For the trajectory dataset, the range of the pedestrian trajectories in the dataset is recorded, and each pedestrian is also regarded as a circle with a radius of 0.3 m. In each test, the center points of the trajectories range's upper and lower boundaries are used as the starting and goal points. A starting time is randomly selected to broadcast pedestrian trajectories based on the dataset and ensure that the pedestrians do not collide with the starting point at the starting time.

In the above scenarios, the robot uses the proposed method, ORCA, LSTM-RL \cite{everett2018motion}, SARL, and ESA methods for planning, respectively. LSTM-RL, SARL, and ESA are all reinforcement-learning-based methods and focus on considering the interaction of the crowd. For each scenario, 300 tests are conducted and the robot's success rate is counted. The test succeeds if the robot reaches the destination, and fails if it collides with the agents, moves out of bounds, or fails to reach the end within the specified time. The LSTM-RL, SARL, and ESA methods are retrained in the same environment as the ESA, and the difference is that the robot is changed from the holonomic robot to the differential robot during the training. When using the proposed method for planning, the navigation path is the connection line between the robot's current and the goal points. The testing results are shown in Tab.~\ref{tab:crowd_env_tests}.

\begin{table}[!t]
    \centering
    \renewcommand{\arraystretch}{1.0}
    \fontsize{8}{8}\selectfont
    \begin{threeparttable}
        \caption{Comparison of different methods in crowd environment tests.} 
        \label{tab:crowd_env_tests}
        \setlength{\tabcolsep}{3pt}
        \begin{tabular}{ccccccc}
        \toprule
        Method & \tabincell{c}{10 agents \\ ORCA} & \tabincell{c}{15 agents \\ ORCA} & \tabincell{c}{20 agents \\ ORCA} & Zara01 & Zara02 & Stu03 \cr
        \midrule
        \midrule
        \addstackgap[4pt] DWA & 0.3$\%$ & 0.3$\%$ & 0$\%$ & 62.3$\%$ & 55$\%$ & 14$\%$ \cr
        \addstackgap[4pt] ORCA & 25.6$\%$ & 11$\%$ & 8.3$\%$ & 76.6$\%$ & 72.6$\%$ & 22.3$\%$ \cr
        \addstackgap[4pt] LSTM-RL & 51.3$\%$ & 34.6$\%$ & 20$\%$ & 60$\%$ & 49.3$\%$ & 1.6$\%$ \cr
        \addstackgap[4pt] SARL & 84.6$\%$ & 61.6$\%$ & 41$\%$ & 69.6$\%$ & 15.3$\%$ & 0$\%$ \cr
        \addstackgap[4pt] ESA & 75.3$\%$ & 56.3$\%$ & 35$\%$ & 67$\%$ & 60$\%$ & 13$\%$ \cr
        \addstackgap[4pt] Ours & \textbf{89.3$\%$} & \textbf{81$\%$} & \textbf{76.6$\%$} & \textbf{92$\%$} & \textbf{93$\%$} & \textbf{68$\%$} \cr
        \bottomrule
        \end{tabular}
    \end{threeparttable}
\end{table}

According to the results in Tab.~\ref{tab:crowd_env_tests}, it can be seen that the proposed method improves the success rate for all the testing scenarios, compared with the current methods. In particular, the proposed method's success rate does not significantly decrease as the environment becomes more complex, which indicates that the proposed method has higher reliability in complex environments. In addition, it can be found that learning-based methods, such as SARL, perform well in the ORCA environment while testing poorly on the pedestrian trajectories dataset. The reason is that these learning-based methods are trained in the ORCA environment, and the data distribution of the pedestrian trajectories dataset and the ORCA environment is quite different, so these methods cannot fit the pedestrian trajectories dataset environment. In contrast, the proposed method does not have the above problems and has better generalization ability.

Furthermore, the examples that use the proposed method to achieve crowd navigation in different environments are shown in Fig.~\ref{fig:crowd_test}. This figure describes the movement process of the robot in crowds using the proposed method. According to Fig.~\ref{fig:crowd_test}(c), the robot first moves to the right to avoid agents and then moves forward for a while. At about 13 seconds, the robot slows down and turns towards the goal point, and it finally arrives at the goal point at 23 seconds.

\subsection{Static Environment Tests}

\begin{figure}[!t]
    \centering
    \includegraphics[width=\linewidth]{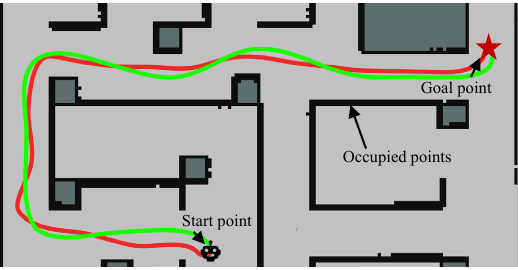}
    \caption{This figure shows an example of the static environment test. The green curve is the robot's moving trajectory using the proposed method and the red curve is the robot's moving trajectory using the TEB method.}
    \label{fig:static_test}
    \vspace{-0.2in}
\end{figure}

In the static environment tests, the testing scenario is shown in Fig.~\ref{fig:static_test}. The start and goal points are randomly selected within the scenario, and the start and goal points are ensured to keep a certain distance from the occupied grid points. The A* algorithm with the Douglas-Peucker algorithm \cite{jian2021global} generates the navigation path. Then, our method and the TEB method are applied to carry out local planning along the navigation path, respectively. Each method conducts 300 tests. Success rate, safety, jitter, time consumption for a single plan, and navigation time from the start point to the goal point are used as comparison metrics. The success rate is measured by the success times that the robot reaches its destination divided by the total testing times. The safety is measured by the minimum distance between the robot and occupied grid points. The jitter is measured by the robot's angular velocity, linear and angular accelerations. Finally, the testing results are shown in Tab.~\ref{tab:static_test}.

\begin{table}[!t]
    \centering
    \renewcommand{\arraystretch}{1.0}
    \fontsize{8}{8}\selectfont
    \begin{threeparttable}
        \caption{Comparison of different methods in static environment tests.} 
        \label{tab:static_test}
        \setlength{\tabcolsep}{1.5pt}
        \begin{tabular}{cccccccc}
        \toprule
        Method & \tabincell{c}{Succ. \\ Rate} & \tabincell{c}{Safety \\ (m)} & \tabincell{c}{Nav. \\ Time \\ (s)} & \tabincell{c}{Mean \\ Ang. vel. \\ (rad/s)} & \tabincell{c}{Mean \\ Lin. acc. \\ (m/$\mathrm{s}^2$)} & \tabincell{c}{Mean \\ Ang. acc. \\ (rad/$\mathrm{s}^2$)} & \tabincell{c}{Time \\ Consuming \\ (ms)} \cr
        \midrule
        \midrule
        \addstackgap[6pt] TEB & 89.6 $\%$ & 0.18 & \textbf{16.14} & 0.46 & 0.44 & 0.65 & \textbf{77.3} $\pm$ 43.8 \cr
        \addstackgap[6pt] Ours & \textbf{97.3} $\%$ & \textbf{0.27} & 18.43 & \textbf{0.44} & \textbf{0.43} & \textbf{0.58} & 94.5 $\pm$ \textbf{11.6} \cr
        \bottomrule
        \end{tabular}
    \end{threeparttable}
\end{table}

It can be seen from Tab.~\ref{tab:static_test} that the proposed method has a 7.7 $\%$ improvement in the success rate, a 50 $\%$ improvement in the safety and a 10.7 $\%$ decline in terms of angular acceleration. This result shows that our method has advantages in safety and jitter, compared with the TEB method. Regarding navigation time, the proposed method spends 14.1 $\%$ more time than the TEB method on average, and part of the reason is that the proposed method tends to choose longer trajectories to keep the distance to the occupied grid points in order to ensure safety. Regarding time consumption for a single plan, the TEB method consumes 18.2 $\%$ time less than the proposed method on average. However, the TEB method's time consumption is unstable. When the number of occupied grid points is large, its time consumption will increase significantly. The TEB method's maximum time consumption can reach 630.1 ms in the experiment, which cannot meet the real-time requirement. In contrast, the proposed method's time consumption is relatively stable, and its maximum time consumption is 146 ms in the experiment, which can fully meet the real-time requirement.

In Fig.~\ref{fig:static_test}, a static environment testing example is shown. It can be clearly seen from the figure that compared with the red curve, the green curve is farther away from the occupied grid points and smoother. This figure demonstrates that compared with the TEB method, the proposed method can effectively improve safety and reduce jitter.

\subsection{Hybrid Environment Demonstration}

\begin{figure}[!t]
    \centering
    \includegraphics[width=\linewidth]{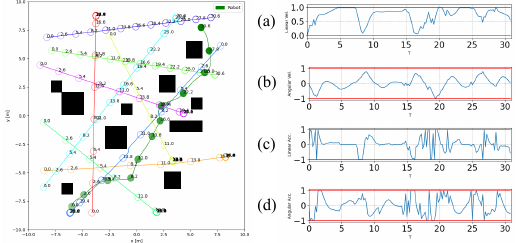}
    \caption{This figure shows a demonstration in the environment with both static and dynamic constraints. The left picture shows the robot's traveling process. The right pictures show the robot's linear velocity, angular velocity, linear acceleration, and angular acceleration during the process.}
    \label{fig:hybrid}
\end{figure}

The proposed method is also tested in an environment with both static and dynamic constraints, as shown in Fig.~\ref{fig:hybrid}. The agents in the environment follow the ORCA policy as Liu \emph{et al.}'s method \cite{liu2020robot}. In the figure, the black areas indicate the obstacles, the colored hollow circles indicate the other agents, and the green circles indicate the robot. It can be seen that the robot successfully avoided the obstacles and agents, and reached the goal point using the proposed method. It can also be seen from the right side of the Fig.~\ref{fig:hybrid} that during this process, the robot's linear velocity, angular velocity, linear acceleration and angular acceleration did not exceed the limitation.

\subsection{Ablation Study}

The scenario used in the ablation study is basically the same as the ORCA environment tests, but the difference is that 25 agents were set up. In this experiment, the difficulty of the test environment is increased to make the comparison of results more distinguishable.

In the ablation study, the robot uses the proposed method without EB-MPC optimization (No Opt.) to plan in the testing scenario. Then, the voxel sampling is further replaced by random sampling (Rand.) for planning. Finally, the traditional distance field similar to \cite{ratliff2009chomp} is used for planning. Each one is tested 300 times, and success rate, jitter, and time consumption for a single plan are recorded. The testing results are shown in Tab.~\ref{tab:ablation_study}.

\begin{table}[!t]
    \centering
    \renewcommand{\arraystretch}{1.0}
    \fontsize{8}{8}\selectfont
    \begin{threeparttable}
        \caption{Comparison of different methods in crowd environment tests.} 
        \label{tab:ablation_study}
        \setlength{\tabcolsep}{4pt}
        \begin{tabular}{cccccc}
        \toprule
        Method & \tabincell{c}{Succ. \\ Rate} & \tabincell{c}{Time \\ Consuming \\ (ms)} & \tabincell{c}{Mean \\ Ang. vel. \\ (rad/s)} & \tabincell{c}{Mean \\ Lin. acc. \\ (m/$\mathrm{s}^2$)} & \tabincell{c}{Mean \\ Ang. acc. \\ (rad/$\mathrm{s}^2$)} \cr
        \midrule
        \midrule
        \addstackgap[4pt] Trad. & 50.6$\%$ & 77.13 & 0.36 & 0.51  & 0.57 \cr
        \addstackgap[4pt] Rand. & 57$\%$ & 76.20 & 0.37 & 0.51 & 0.57 \cr
        \addstackgap[4pt] No Opt. & \textbf{67.6$\%$} & \textbf{73.03} & 0.41 & 0.57 & 0.70 \cr
        \addstackgap[4pt] Complete & 67.3$\%$ & 103.53 & \textbf{0.35} & \textbf{0.29} & \textbf{0.38} \cr
        \bottomrule
        \end{tabular}
    \end{threeparttable}
        \vspace{-0.2in}

\end{table}

According to the testing results, the success rate is increased by $6.4\%$, when the proposed time-varying distance fields are used instead of the traditional distance field. The voxel sampling also improves $10.6 \%$ in the success rate than the random sampling. This result can prove the effectiveness of both in complex crowd environments. Furthermore, the robot's jitter can be significantly reduced almost without reducing the success rate when adding the optimization method. Especially in linear and angular accelerations, there are reductions of $49.1 \%$ and $45.7 \%$, respectively. In terms of time consumption, replacing the distance fields and using voxel sampling hardly caused an increase in time consumption. Although the optimization method increases the time consumption by $41.7 \%$, it can still guarantee the real-time performance of the method. In conclusion, all parts of the proposed method are proved effective as expectation.

\section{CONCLUSIONS}

This paper proposes a long-term dynamic window approach local planning method for differential wheeled robots. This method can be applied to both crowd and static environments, and its planned state sequence in real time can ensure the safety and reduce jitter of the robot while satisfying the kinodynamic constraints. The limitation of the method is that it does not consider the interaction between the crowd or the interaction between the robot and the crowd, which is the limitation of our method. In future work, we will consider integrating the prediction of other agents\cite{sun2019interactive, sun2022pseudo} the interaction between the crowd in the planning to further improve the performance.






\bibliographystyle{IEEEtran}
\bibliography{reference}

\end{document}